\documentclass[10pt, a4paper]{article}
\usepackage{lrec2000}
\usepackage{graphicx}

\title{Corpus based Enrichment of GermaNet Verb Frames}

\name{Manuela Kunze and Dietmar R{\"o}sner}

\address{ Otto-von-Guericke-Universit{\"a}t Magdeburg\\
Institut f{\"u}r Wissens- und Sprachverarbeitung \\
P.O. Box 4120, D--39016 Magdeburg,\\ Germany \\
               makunze, roesner@iws.cs.uni-magdeburg.de}

\abstract{Lexical semantic resources, like WordNet, are often used
in real applications of natural language document processing. For
example, we integrated GermaNet in our document suite XDOC. In
addition to hypernymy and synonymy relations, we want to exploit
GermaNet verb frames for our analysis. In this paper, we outline
an approach for the domain related enrichment of GermaNet verb
frames by corpus based syntactic and co-occurrence data analyses
of real documents.}

\begin{document}

\maketitleabstract

\hyphenation{Verb-valen-zen Ver-kehrs-schild}
\section{Introduction}
Lexical resources, like WordNet \cite{fellbaum:98}, GermaNet
\cite{hamp.feldweg:1997,kunze:2001}, or EuroWordNet, will be more
than ever applied as resources in applications, like Text Mining
and Data Mining. Often it is necessary to extend or to adapt the
resources for the application (see also
\cite{vossen:2001,navigli.velardi:2002}).

In \cite{kunze.roesner:2003germanet} we presented the integration
of the lexical resource GermaNet into our document processing
system XDOC. In XDOC, GermaNet is used as linguistic (lexical)
resource for tasks like
\begin{itemize}
    \item semantic tagging of tokens,
    \item case frame analysis, and
    \item semantic interpretation of syntactic structures (SIsS).
\end{itemize}
One problem for the integration of GermaNet resources was the
usage of GermaNet's verb frames. The problem there was that the
information encoded in GermaNet verb frames is not sufficient
(i.e., not detailed enough) for usage within the case frame
analysis of XDOC.

Case frame analysis needs detailed information about the syntactic
form (e.g., which preposition and which case of the PP), the
semantic category of the filler of a relation, and which thematic
role is described by the relation (e.g., agent, location, etc.)
for a complement of a verb.

GermaNet's verb frames have two deficits with respect to usage in
XDOC:
\begin{enumerate}
    \item for verbs, the information given is incomplete (e.g.,
    preposition, semantic category, and thematic role are missing) and
    \item for nouns, no frame information is available.
\end{enumerate}

The information missing can be classified into several types:
lexical (preposition), syntactic (case of noun phrase in a
preposition phrase\footnote{For example, the preposition \emph{in}
can required a NP with the case \emph{accusative} or
\emph{dative}.}) and semantic (category of the filler and the
thematic role of the relation described). The creation or manual
adaption of GermaNet's resources is time-consuming. Related works
included the automatic building of subcategorisation lexicons for
German verbs \cite{wauschkuhn99,SchulteImWalde:02a} and the
automatic identification of thematic roles (see
\cite{gildea.hockenmaier:2003}; \cite{gildea.jurafsky:2002}) by
exploitation of a syntactically parsed corpus as input data. The
method described in this paper extracts the necessary syntactic
information and information about the required semantic categories
for possible complements of a verb. This approach uses a corpus
annotated with chunks (noun phrases and prepositional phrases) and
GermaNet's verb frames information.

This paper is organized as follows: first we give a short
description about the evaluation environment. After this, we
present the methodology by discussing an example. This is followed
by the presentation and discussion of results from our
experiments.

\section{Evaluation Environment}
\subsection{Evaluation Corpus}
For our work we used a corpus of medical documents in German
(forensic autopsy protocols) with more than 1 million running word
forms. The autopsy protocols have a strictly defined content and
layout. They are separated into different document parts, e.g.
findings, background, discussion, death causes, etc. Each document
part has its own characteristics (sub-language).

The analyses with XDOC are concentrated on the sections of
\emph{findings}, \emph{background} and \emph{discussion}. The
\emph{findings} section contains a high ratio of nouns and
adjectives and syntactic (sentence) structures are mostly short.
This section describes the medical findings in an everyday
vocabulary without domain specific (medical) terms. A standard
distribution of all word classes and regular syntactic structures
occurs in the \emph{background} and \emph{discussion} sections.
The \emph{background} section describes, for example, the details
of a traffic accident, while the section \emph{discussion}
contains a combination of the results of the \emph{finding}
section and the facts reported in the \emph{background} section.
Both document parts contains a high and multifaceted number of
named entities (NE). For example, each forensic autopsy protocol
has a registration number (e.g., G 123/45), which is often
referred to in the document. Furthermore, other NEs like date
specifications, names of locations (e.g., streets, cities), or
names of persons etc., occur in the texts.

The analysis described in this paper is only concentrated on the
document parts \emph{background} and \emph{discussion}. These
parts were chosen, because both document parts contain regular
syntactic structures of German and a minor ratio of domain
specific terms. The tokens seem to belong much more likely to
everyday language than to a sub-language of a specific domain.

\subsection{Tools and Resources}
For the adaptation of GermaNet verb frames, a syntactically
annotated corpus is required. For this and other preprocessing
steps, the document suite XDOC\footnote{For a full description of
the methods inside XDOC see \cite{roesner.kunze:2002coling}.} is
used. In particular, following preprocessing steps of XDOC are
used:
\begin{itemize}
    \item sentence splitter,
    \item POS tagger,
    \item syntactic parser.
\end{itemize}
These methods output their results as XML structures, which are
accepted by the subsequent processing steps based on XML
structures as their input. For the extraction of relevant
information (syntactic structures), XSL transformation is applied
\cite{clark:1999}.

\begin{figure}[hbtp]
\begin{center}
\includegraphics[scale=0.5]{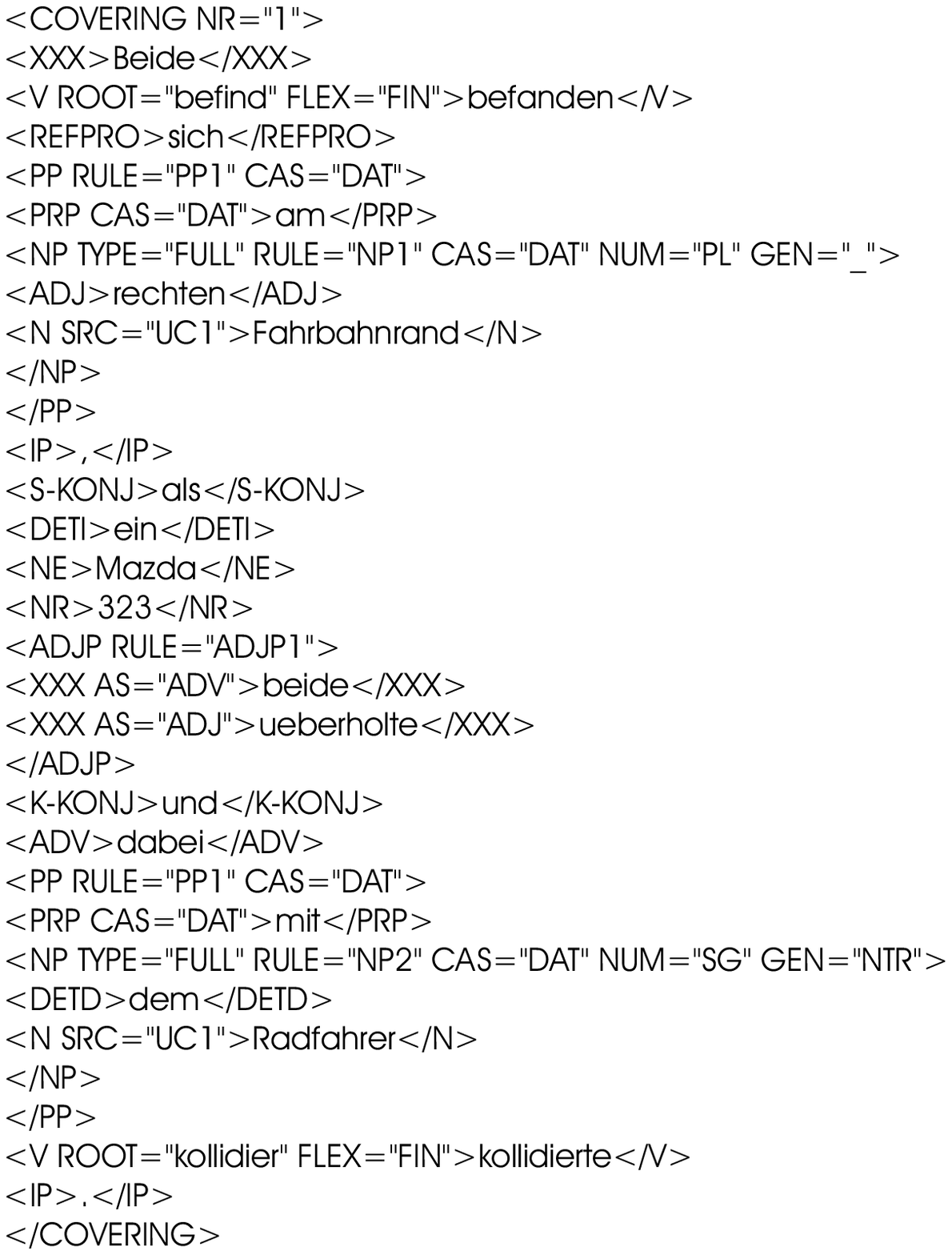}
\caption{A syntactically parsed sentence with NPs and PPs
chunks.}\label{idea}
\end{center}
\end{figure}

The quality of expected results is strongly dependent on the
quality of input data, especially the results of the chart parser.
The syntactic parser of XDOC is a bottom-up chart parser, which
works with a context free grammar for German (ca. 400 rules). The
robust XDOC syntactic parser outputs sentences completely parsed
(readings) or only structures partially parsed (coverings). In
these coverings, basic structures, like noun phrases or
prepositional phrases (frequent elements in GermanNet verb frames)
are annotated by XDOC's parser (see Fig. \ref{idea}).

\section{An Outline of the Approach}
The basic assumption for the approach is that in a corpus with
similar texts (news, expert's report, abstracts, etc.) a frequent
verb co-occurs with the same complements. The complements of such
a verb often appear in a similar syntactic structure at the same
position in a sentence. Further, the fillers have the same
semantic category. In the case of the autopsy corpus the number of
authors of the documents is small. This results in a high rate of
repetition of specific wordings or phrases, because authors have
the tendency to use the same phrase for the description of similar
facts (author style).

The steps of the procedure are:
\begin{itemize}
    \item use the verb frames given by GermaNet as simple patterns
    for the recognition of potential candidates for case frames in the corpus,
    \item extract information about prepositions used in a complement (element of the
case frame candidate), and
    \item count the occurrences of similar semantic fillers (roles) for an
    element of the case frame.
\end{itemize}

The approach is presented by considering the verbs:
\emph{verstarb} (to pass away), \emph{kollidieren} (to collide),
\emph{befahren} (to cruise), \emph{operieren} (to operate) as
examples.

\begin{table}[ht]
 \begin{center}
 \small
 \begin{tabular}{|l|l|l|}
       \hline
       verb& occurrences & frame information from GermaNet\\
       \hline\hline
       kollidieren & 34 & NN.Pp\\
       operieren &  14 &NN.AN or NN.AN.BL\\
       versterben & 128 &NN.BT  \\
       erfolgen & 187 &NE.AN or NN.PP  \\
       befahren & 59 &NN.AN or NN.AN.AZ or NN.AN.BM\\
       ereignen & 29 & NE or NE.AR or NN.AR.BT or NN.BL\\
       \hline
 \end{tabular}
 \caption{Verbs and their GermaNet verb frame information.}
 \label{tab-verbs}
 \end{center}
 \end{table}
 \normalsize

All these verbs have only one sense in GermaNet with the correct
meaning for our cases. But multiple verb frames can be assigned to
a sense, see for example the verb \emph{befahren} (see also table
\ref{tab-verbs})\footnote{A detailed description of the notation
of the verb frames is available at:
http://www.sfs.nphil.uni-tuebingen.de/lsd/} with 3 verb frames.

GermaNet verb frames characterise the syntactic subcategorisation
of a verb. The elements of a verb frame describe different
complements of verbs. For example, '\emph{NN}' or '\emph{AN}'
stand for a noun phrase in case \emph{nominative} resp.
\emph{accusative}, '\emph{PP}' represents a prepositional phrase
but without information about the concrete preposition used. Both
elements give no information about the semantic category of the
role filler and the thematic role. Only elements like, '\emph{BM}'
or '\emph{BL}' (stands for an adverbial complement or a
prepositional phrase, which indicates a \emph{manner} or
\emph{local} complement), give some semantic restriction for the
filler.

In the following, we do sketch the approach: The initial basis is
a corpus of documents, which are separated into a sequence of
sentences.

To complete the case frame analysis of a frequent verb in the
corpus, all sentences in which the verb occurs were selected.
These sentences are annotated by the POS Tagger of XDOC and are
parsed by the syntactic parser of XDOC. For the analysis, only the
annotation of NPs and PPs is required. In this case, the grammar
of the chart parser was reduced to 15 rules for the annotation of
basic structures, like noun phrases and prepositional phrases.

According to the verb frame information of GermaNet, possible
candidates are selected from the structures parsed by the usage of
XSL transformations. For example, for elements like \emph{NN} and
\emph{AN}, noun phrases with the case nominative or accusative
resp. are selected. Elements like \emph{PP} \footnote{The GermaNet
notation 'PP' means a required prepositional phrase.} are
prepositional phrases with non specified case or preposition.
Other elements are ambiguous, like the element \emph{BM}. The
syntactic realisation of \emph{BM} could be a prepositional phrase
or an adverb. In this case, both realizations must be considered
during search and analysis.

The GermaNet verb frame of the verb \emph{collide} contains the
following information: 'NN.Pp' -- a noun phrase in case
\emph{nominative} and an \emph{optional} prepositional phrase.
Following sentences occur in the corpus:

{\begin{itemize}
    \item \emph{Der erste H\"anger kollidierte vermutlich mit der vorderen rechten
Seite mit einem ... Haus.}
    \item \emph{... sein LKW kollidierte mit dem PKW.}
    \item \emph{Der Pkw ... kollidierte mit 3 Begrenzungsst\"aben.}
    \item \emph{Der ... Pkw Peugeot hingegen kollidierte frontal mit dem Pkw
Renault}
    \item \emph{Nachfolgend kollidierten 3 Pkw mit dem VW Golf.}

\end{itemize}

The first assignment ('NN') is easy to handle, all noun phrases
with the case \emph{nominative} are selected. The following
instances can be assigned to the 'NN' element of the verb frame
given the sentences above:

\begin{itemize}
    \item \emph{NN:} \emph{der erste H\"anger}, \emph{sein LKW}, \emph{der Pkw}, \emph{der Pkw Peugeot}, \emph{3 Pkw}
\end{itemize}

Semantic classification uses information available in GermaNet. In
the first step, GermaNet top level information is used as a
shallow classification. To improve the classification, the
hypernymy tree information of GermaNet is exploited.

For the verb \emph{collide}, following two types of fillers for
the 'NN' element are encountered in the corpus. The first type is
a person referenced by pronouns (11), registration numbers (6),
like \emph{G 1234/11}, or by a person name (1). And the second
type of the filler is a vehicle (16). Both types describe traffic
participants (road users).

In the next step, the occurrences of prepositions in these
sentences are counted. The verb \emph{collide} occurs in 34
sentences within the corpus. The frequent prepositions used in
these sentences are presented in table \ref{tab-prep}.

\begin{table}[ht]
 \begin{center}
 \small
  \begin{tabular}{|l|l|l|}
       \hline
       preposition & ratio & semantic\\
       \hline\hline
       mit (dat) & 54 & temporal, instrument, modal, causal\\
       auf (dat, acc) &  23 & local, temporal, modal, causal\\
       aus (dat)& 17 & local, causal\\
       am (dat) & 17 & modal, temporal\\
       nach (dat) & 15 & local, temporal, final, modal \\
       in (dat, acc) & 13 & local, temporal, modal\\
       von (dat) & 11 & local, temporal, modal\\
       \hline
 \end{tabular}
 \caption{High frequent prepositions co-occurred with the verb \emph{collide}.}
  \label{tab-prep}
  \end{center}
 \end{table}
\normalsize

Table \ref{tab-prep} shows, that different prepositions are
possible as the indicator for the prepositional phrase of the verb
frame. For the selection of the correct preposition, following
assumptions are used. The PPs co-occurring with instances of the
verb \emph{kollidieren} (collide) are evaluated. Most prepositions
allow for different semantic interpretation (cf. \ref{tab-prep}).
Disambiguation is only possible when taking the classification of
the embedded NP into account. Only PPs that are not referring to
local or temporal circumstances are counted, because temporal or
local adjuncts can co-occur with most verbs.\footnote{The treament
is different if the GermaNet patterns contain explicit elements
about locale or temporal information, like \emph{BT} or
\emph{BL}.} From the remaining PPs the preposition with the
highest frequency is taken as candidate for the derived case
frame.

%On the one hand, high frequent prepositions are prefered, and on
%the other hand, prepositions, which only describe locale or
%temporal attributes\footnote{The treament is different if  the
%GermaNet patterns contain explicit elements about locale or
%temporal information, like \emph{BT} or \emph{BL}.} are ignored.
%We assume that these attributes are not explicitly described in a
%GermaNet verb frame, because these elements can co-occur with most
%verbs.

Furthermore, this approach can be enhanced, when the information
about the distance between verb and potential prepositional phrase
complement is exploited. In a sentence, it is possible that the
same preposition can occur more than once in a PP. Coordination is
one of example:
\begin{itemize}
    \item \emph{Nach Angaben der anwesenden Kliniker soll er \textbf{mit} einem PKW
von der Fahrbahn abgekommen sein und dort \textbf{mit}
feststehenden Gegenst{\"a}nden kollidiert sein.}
\end{itemize}

For examples like this, the approach described can be enhanced in
the following way: Only clause structures instead of whole
sentences are explored, because more than one verb occur in a
sentence. The clauses in these sentences are splitted by commas or
by the conjunction '\emph{und}'(and). In addition, a heuristic is
used: Only PPs situated next to the verb, before or after the verb
(scope of a verb), are analysed. Further work will be the
refinement of this simple heurisics.

For the verb \emph{collide} the following prepositions were
encountered as results: 27 times 'mit', 5 times 'nach', twice the
prepositions 'beim', 'am', 'als, and once the preposition 'in'.

The filler of the 'Pp' with the preposition 'mit' can be assigned
to the semantic category 'solid object'.

\begin{itemize}
    \item \emph{Pp:} \emph{mit einem Pkw}, \emph{mit einem Baum}, \emph{mit dem
    Mercedes}, \emph{mit der Mittelleitplanke}, \emph{mit einem Verkehrsschild}
\end{itemize}

In sum, the approach results in the following extended verb frame
of the verb \emph{collide}: \begin{itemize}
    \item The filler of 'NN' element can be either a person (e.g., NE, pronoun) or
    a vehicle (e.g., a 'regular' noun, like \emph{PKW}).
    \item The 'Pp' element describes an object in the syntactic
    form of a PP with the preposition 'mit' and the case
    \emph{dative}. In this case, the semantic category of the filler is the category  \emph{solid object}.
\end{itemize}

For the verb \emph{befahren} there exist three verb frames in
GermaNet, each consisting of the elements \emph{NN} and \emph{AN}.
For these elements, the approach described above delivers
following details:
\begin{itemize}
    \item \emph{NN}: The results contain here again instances of traffic participants. The first
    subcategory describes persons, in form of NEs (registration number or name
    of the person), pronouns, or with the noun 'driver' in phrases like \emph{driver of the 'car'}.
    The second subcategory is presented through
    vehicles, like NEs (\emph{PKW VW Lupo}) or as nouns, like car, tramway, motor vessel etc.
    \item \emph{AN}: All possible candidates (e.g., street, German freeway, avenue, canal, etc.), which are found in GermaNet, could be assigned to \emph{traffic
    route}.
\end{itemize}
The additional elements \emph{AZ}\footnote{\emph{zu-infinitive}
complement} and \emph{BM} were not analysed, because up to now
this work was restricted to the enrichment of elements, like noun
and prepositional phrases. The extension to other possible
elements in a verb frame is part of our future work.

\section{Discussion}
The results obtained via this approach can support a designer for
verb frames. Based on the verb frame information in GermaNet, this
method delivers possible candidates of fillers for noun and
prepositional phrases . The results contains information about the
semantic category and syntactic form (and elements, like
prepositions) of case roles fillers.

The results are dependent on a good lexical coverage to get the
correct semantic information for a filler and strongly dependent
on the correct annotation of syntactic structures. The number of
the coverings delivered for a sentence by the parser can be
reduced. At first only coverings are allowed, which are in
accordance with GermaNet verb frames. Second, only relevant parts
(clause) of a complex sentence are extracted.

One problem, which occurs was the correct handling of NEs in the
corpus. In addition to date or time information, the approach must cover
names of locations (e.g., streets, like \emph{A 9}), names of
vehicles (e.g., \emph{Opel Frontera}), names of persons (e.g.,
\emph{Mr. Miller'}), and registration numbers (e.g., of persons:
\emph{G 1345/78}; or licence plate numbers: \emph{ABZ AB-789}).
%In
%most cases, these data were to make anonymous by different
%authors. So, the method used here must handle with different anonymous forms of the
%same token, for example \emph{XY-Strasse}, \emph{X-Strasse}, or
%\emph{Strasse X}.

\section{Conclusion} In this paper, we described an approach for
the enrichment of GermaNet's verb frames. It is based on
co-occurrence data analyses of a corpus of forensic autopsy
protocols.

%The German document suite XDOC and XSL transformations were used
%to get additional information about the verb frames. The
%evaluation corpus was a corpus of forensic autopsy protocols, but
%the verb frames are assigned to more everyday verbs than to domain
%specific verbs.

For the approach described above, the document parts
\emph{background} and \emph{discussion} from the forensic autopsy
protocol were used. These parts were chosen, because in these
parts, a minor ratio of domain specific terms occurs. The form and
the content of the \emph{background} section are similar to a
report in a newspaper.

The approach outlined here is based on structural and syntactical
analysis and on the analysis of co-occurrence data. These
co-occurrence data were verbs and syntactical structures in the
neighborhood of the verbs. The quality of the results are strongly
dependent on the results of the syntactic parser and the correct
handling of named entities. Both can be enhanced by an improvement
of the domain specific resources, like the grammar of the chart
parser. Our future work will be to confirm and to evaluate the
approach with another corpus, for example the EUROPARL corpus --
available at http://www.isi.edu/$\sim$koehn.

\bibliographystyle{lrec2000}
\bibliography{lrec2004}

\begin{thebibliography}{12}
\expandafter\ifx\csname natexlab\endcsname\relax\def\natexlab#1{#1}\fi

\bibitem[Clark, 1999]{clark:1999}
Clark, J., 1999.
\newblock {XSL} {T}ransformations {(XSLT)} {V}ersion 1.0.
\newblock W3c recommendation, {World Wide Web Consortium}.
\newblock URL:http//www.w3.org/TR/xslt.

\bibitem[Fellbaum, 1998]{fellbaum:98}
Fellbaum, C., 1998.
\newblock {\em {W}ord{N}et: {A}n {E}lectronic {L}exical {D}atabase\/}.
\newblock Mass.: MIT Press.

\bibitem[Gildea and Hockenmaier, 2003]{gildea.hockenmaier:2003}
Gildea, Daniel and Julia Hockenmaier, 2003.
\newblock Identifying semantic roles using combinatory categorial grammar.
\newblock In {\em 2003 Conference on Empirical Methods in Natural Language
  Processing (EMNLP)\/}. Sapporo, Japan.

\bibitem[Gildea and Jurafsky, 2002]{gildea.jurafsky:2002}
Gildea, Daniel and Daniel Jurafsky, 2002.
\newblock Automatic labeling of semantic roles.
\newblock {\em Computational Linguistics\/}, 28(3):245--288.

\bibitem[Hamp and Feldweg, 1997]{hamp.feldweg:1997}
Hamp, B. and H.~Feldweg, 1997.
\newblock {G}erma{N}et -- a lexical-semantic {N}et for {G}erman.
\newblock In P.~Vossen et.al. (ed.), {\em {Proc. of ACL/EACL-97 workshop
  Automatic Information Extraction and Building of Lexical Semantic Resources
  for NLP Applications}\/}. Madrid.

\bibitem[Kunze, 2001]{kunze:2001}
Kunze, C., 2001.
\newblock {\em {L}exikalisch-semantische {W}ortnetze\/}.
\newblock Heidelberg; Berlin: Spektrum, Akademischer Verlag, pages 386--393.

\bibitem[Kunze and R{\"o}sner, 2003]{kunze.roesner:2003germanet}
Kunze, M. and D.~R{\"o}sner, 2003.
\newblock {I}ssues in {E}xploiting {GermaNet} as a {R}esource in {R}eal
  {A}pplications.
\newblock In {\em {GermaNet-Workshop: Anwendungen des deutschen Wortnetzes in
  Theorie und Praxis}\/}. T{\"u}bingen, Germany.

\bibitem[Navigli and Velardi, 2002]{navigli.velardi:2002}
Navigli, R. and P.~Velardi, 2002.
\newblock {A}utomatic {A}daption of {W}ord{N}et to {D}omains.
\newblock In {\em {Proc. of LREC 2002}\/}. Las Palmas.

\bibitem[R{\"o}sner and Kunze, 2002]{roesner.kunze:2002coling}
R{\"o}sner, D. and M.~Kunze, 2002.
\newblock {A}n {XML} {B}ased {D}ocument {S}uite.
\newblock In {\em Coling 2002\/}. Taipei, Taiwan.

\bibitem[{Schulte im Walde}, 2002]{SchulteImWalde:02a}
{Schulte im Walde}, Sabine, 2002.
\newblock A {S}ubcategorisation {L}exicon for {G}erman {V}erbs induced from a
  {L}exicalised {PCFG}.
\newblock In {\em Proceedings of the 3rd Conference on Language Resources and
  Evaluation\/}, volume~IV. Las Palmas de Gran Canaria, Spain.

\bibitem[Vossen, 2001]{vossen:2001}
Vossen, P., 2001.
\newblock {E}xtending, {T}rimming and {F}using {W}ord{N}et for technical
  {D}ocuments.
\newblock In {\em {Proc. of NAACL 2001 workshop on WordNet and Other Lexical
  Resources}\/}. Pittsbourgh.

\bibitem[Wauschkuhn, 1999]{wauschkuhn99}
Wauschkuhn, O., 1999.
\newblock {\em Automatische Extraktion von Verbvalenzen aus deutschen
  Textkorpora\/}.
\newblock Ph.D. thesis, Institut f{\"u}r Informatik, Universit{\"a}t Stuttgart.

\end{thebibliography}
\end{document}